# Gene Expression Programming: A New Adaptive Algorithm for Solving Problems

**Cândida Ferreira**[†]

Departamento de Ciências Agrárias
Universidade dos Açores
9701-851 Terra-Chã
Angra do Heroísmo, Portugal

Gene expression programming, a genotype/phenotype genetic algorithm (linear and ramified), is presented here for the first time as a new technique for the creation of computer programs. Gene expression programming uses character linear chromosomes composed of genes structurally organized in a head and a tail. The chromosomes function as a genome and are subjected to modification by means of mutation, transposition, root transposition, gene transposition, gene recombination, and one- and two-point recombination. The chromosomes encode expression trees which are the object of selection. The creation of these separate entities (genome and expression tree) with distinct functions allows the algorithm to perform with high efficiency that greatly surpasses existing adaptive techniques. The suite of problems chosen to illustrate the power and versatility of gene expression programming includes symbolic regression, sequence induction with and without constant creation, block stacking, cellular automata rules for the density-classification problem, and two problems of boolean concept learning: the 11-multiplexer and the GP rule problem.

## 1. Introduction

Gene expression programming (GEP) is, like genetic algorithms (GAs) and genetic programming (GP), a genetic algorithm as it uses populations of individuals, selects them according to fitness, and introduces genetic variation using one or more genetic operators [1]. The fundamental difference between the three algorithms resides in the nature of the individuals: in GAs the individuals are linear strings of fixed length (chromosomes); in GP the individuals are nonlinear entities of different sizes and shapes (parse trees); and in GEP the individuals are encoded as linear strings of fixed length (the genome or chromosomes) which are afterwards expressed as nonlinear entities of different sizes and shapes (i.e., simple diagram representations or expression trees).

If we have in mind the history of life on Earth (e.g., [2]), we can see that the difference between GAs and GP is only superficial: both systems use only one kind of entity which functions both as genome and body (phenome). These kinds of systems are condemned to have one of two limitations: if they are easy to manipulate genetically, they lose in functional complexity (the case of GAs); if they exhibit a certain amount of functional complexity, they are extremely difficult to reproduce with modification (the case of GP).

In his book, *River Out of Eden* [3], R. Dawkins gives a list of thresholds of any life explosion. The first is the *replicator threshold* which consists of a self-copying system in which there is hereditary variation. Also important is that replicators survive by virtue of their own properties. The second threshold is the *phenotype threshold* in which replicators survive by virtue of causal effects on something else - the phenotype. A simple example of a replicator/phenotype system is the DNA/protein system of life on Earth. For life to move beyond a very rudimentary stage, the phenotype threshold should be crossed [2, 3].

Similarly, the entities of both GAs and GP (simple replicators) survive by virtue of their own properties. Understandingly, there has been an effort in recent years by the scientific community to cross the phenotype threshold in evolutionary computation. The most prominent effort is developmental genetic programming (DGP) [4] where binary strings are used to encode mathematical expressions. The expressions are decoded using a five-bit binary code, called *genetic code*. Contrary to its analogous natural genetic code, this "genetic code", when applied to binary strings, frequently produces invalid expressions (in nature there is no such thing as an invalid protein). Therefore a huge amount of computational resources goes toward editing these illegal structures, which limits this system considerably. Not surprisingly, the gain in performance of DGP over GP is minimal [4, 5].

---

[†] Electronic mail and web addresses: candidaf@gene-expression-programming.com; http://www.gene-expression-programming.com.

Present address: Gepsoft, 37 The Ridings, Bristol BS13 8NU, UK.

The interplay of chromosomes (replicators) and expression trees (phenotype) in GEP implies an unequivocal translation system for translating the language of chromosomes into the language of expression trees (ETs). The structural organization of GEP chromosomes presented in this work allows a truly functional genotype/phenotype relationship, as any modification made in the genome always results in syntactically correct ETs or programs. Indeed, the varied set of genetic operators developed to introduce genetic diversity in GEP populations always produces valid ETs. Thus, GEP is an artificial life system, well established beyond the replicator threshold, capable of adaptation and evolution.

The advantages of a system like GEP are clear from nature, but the most important should be emphasized. First, the chromosomes are simple entities: linear, compact, relatively small, easy to manipulate genetically (replicate, mutate, recombine, transpose, etc.). Second, the ETs are exclusively the expression of their respective chromosomes; they are the entities upon which selection acts and, according to fitness, they are selected to reproduce with modification. During reproduction it is the chromosomes of the individuals, not the ETs, which are reproduced with modification and transmitted to the next generation.

On account of these characteristics, GEP is extremely versatile and greatly surpasses the existing evolutionary techniques. Indeed, in the most complex problem presented in this work, the evolution of cellular automata rules for the density-classification task, GEP surpasses GP by more than four orders of magnitude.

The present work shows the structural and functional organization of GEP chromosomes; how the language of the chromosomes is translated into the language of the ETs; how the chromosomes function as genotype and the ETs as phenotype; and how an individual program is created, matured, and reproduced, leaving offspring with new properties, thus, capable of adaptation. The paper proceeds with a detailed description of GEP and the illustration of this technique with six examples chosen from different fields.

## 2. An overview of gene expression algorithms

The flowchart of a gene expression algorithm (GEA) is shown in Figure 1. The process begins with the random generation of the chromosomes of the initial population. Then the chromosomes are expressed and the fitness of each individual is evaluated. The individuals are then selected according to fitness to reproduce with modification, leaving progeny with new traits. The individuals of this new generation are, in their turn, subjected to the same developmental process: expression of the genomes, confrontation of the selection environment, and reproduction with modification. The process is repeated for a certain number of generations or until a solution has been found.

Note that reproduction includes not only replication but also the action of genetic operators capable of creating genetic diversity. During replication, the genome is copied and transmitted to the next generation. Obviously, replication alone cannot introduce variation: only with the action of the remaining operators is genetic variation introduced into the population. These operators randomly select the chromosomes to be modified. Thus, in GEP, a chromosome might be modified by one or several operators at a time or not be modified at all. The details of the implementation of GEP operators are shown in section 5.

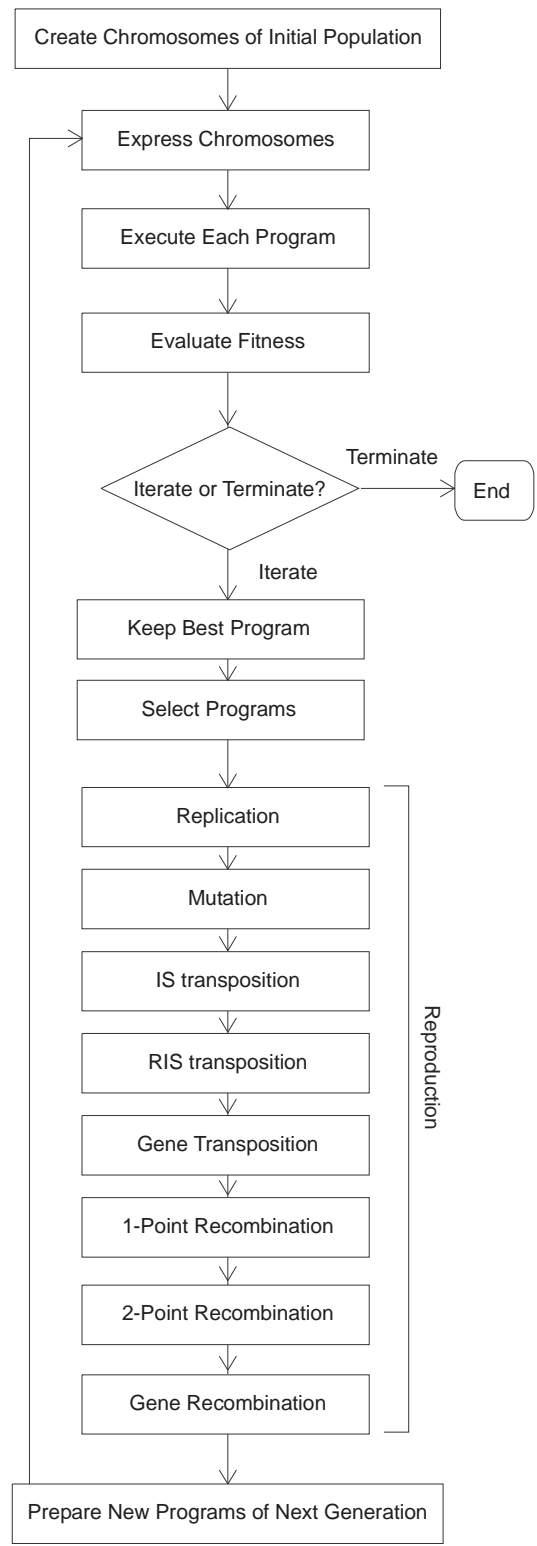

**Figure 1.** The flowchart of a gene expression algorithm.

## 3. The genome of gene expression programming individuals

In GEP, the genome or chromosome consists of a linear, symbolic string of fixed length composed of one or more genes. It will be shown that despite their fixed length, GEP chromosomes can code ETs with different sizes and shapes.

### 3.1. Open reading frames and genes

The structural organization of GEP genes is better understood in terms of open reading frames (ORFs). In biology, an ORF, or coding sequence of a gene, begins with the "start" codon, continues with the amino acid codons, and ends at a termination codon. However, a gene is more than the respective ORF, with sequences upstream from the start codon and sequences downstream from the stop codon. Although in GEP the start site is always the first position of a gene, the termination point does not always coincide with the last position of a gene. It is common for GEP genes to have noncoding regions downstream from the termination point. (For now we will not consider these noncoding regions, because they do not interfere with the product of expression.)

Consider, for example, the algebraic expression:

$$\sqrt{(a+b)\times(c-d)}, \quad (3.1)$$

which can also be represented as a diagram or ET:

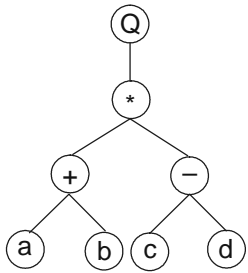

where "Q" represents the square root function. This kind of diagram representation is in fact the phenotype of GEP individuals, being the genotype easily inferred from the phenotype as follows:

```
01234567
Q*+-abcd                     (3.2)
```

which is the straightforward reading of the ET from left to right and from top to bottom. Expression (3.2) is an ORF, starting at "Q" (position 0) and terminating at "d" (position 7). These ORFs were named K-expressions (from the Karva language, the name I chose for the language of GEP). Note that this ordering differs from both the postfix and prefix expressions used in different GP implementations with arrays or stacks [6].

The inverse process, that is, the translation of a K-expression into an ET, is also very simple. Consider the following K-expression:

```
01234567890
Q*+*a*Qaaba                  (3.3)
```

The start position (position 0) in the ORF corresponds to the root of the ET. Then, below each function are attached as many branches as there are arguments to that function. The assemblage is complete when a baseline composed only of terminals (the variables or constants used in a problem) is formed. In this case, the following ET is formed:

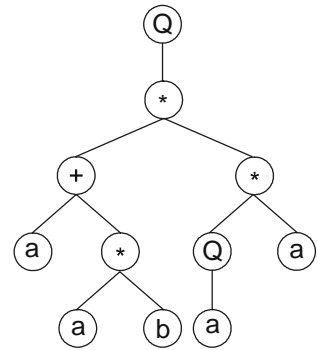

Looking only at the structure of GEP ORFs, it is difficult or even impossible to see the advantages of such a representation, except perhaps for its simplicity and elegance. However, when ORFs are analyzed in the context of a gene, the advantages of such representation become obvious. As stated previously, GEP chromosomes have fixed length and are composed of one or more genes of equal length. Therefore the length of a gene is also fixed. Thus, in GEP, what varies is not the length of genes (which is constant), but the length of the ORFs. Indeed, the length of an ORF may be equal to or less than the length of the gene. In the first case, the termination point coincides with the end of the gene, and in the second case, the termination point is somewhere upstream from the end of the gene.

So, what is the function of these noncoding regions in GEP genes? They are, in fact, the essence of GEP and evolvability, for they allow modification of the genome using any genetic operator without restrictions, always producing syntactically correct programs without the need for a complicated editing process or highly constrained ways of implementing genetic operators. Indeed, this is the paramount difference between GEP and previous GP implementations, with or without linear genomes (for a review on GP with linear genomes see [7]).

### 3.2. Gene expression programming genes

GEP genes are composed of a head and a tail. The head contains symbols that represent both functions (elements from the function set *F*) and terminals (elements from the terminal set *T*), whereas the tail contains only terminals. Therefore two different alphabets occur at different regions within a gene. For each problem, the length of the head $h$ is chosen, whereas the length of the tail $t$ is a function of $h$ and the number of arguments of the function with the most arguments $n$, and is evaluated by the equation

$$t = h(n-1) + 1. \qquad (3.4)$$

Consider a gene composed of {Q, *, /, -, +, a, b}. In this case $n = 2$. For instance, for $h = 10$ and $t = 11$, the length of the gene is 10+11=21. One such gene is shown below (the tail is shown in bold):

```
0123456789012345 67890
+Q-/b*aaQbaabaabbaaab           (3.5)
```

and it codes for the following ET:

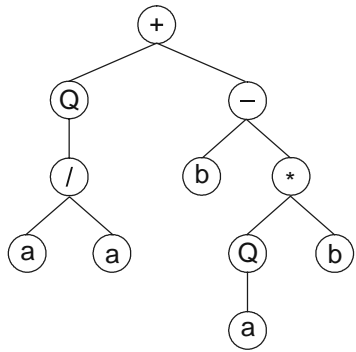

In this case, the ORF ends at position 10, whereas the gene ends at position 20.

Suppose now a mutation occurred at position 9, changing the "b" into "+". Then the following gene is obtained:

```
0123456789012345 67890
+Q-/b*aaQ+aabaabbaaab           (3.6)
```

and its ET gives:

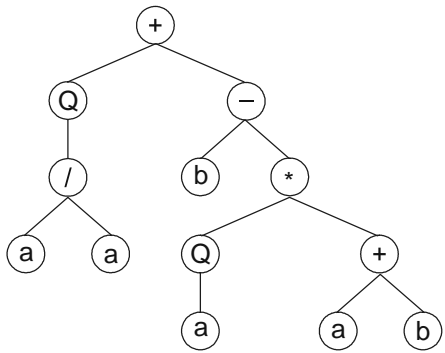

In this case, the termination point shifts two positions to the right (position 12).

Suppose now that a more radical modification occurred, and the symbols at positions 6 and 7 in gene (3.5) change respectively into "+" and "*", creating the following gene:

```
0123456789012345 67890
+Q-/b*+*Qbaabaabbaaab           (3.7)
```

giving the ET:

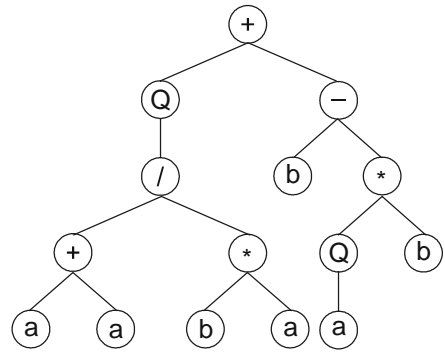

In this case the termination point shifts several positions to the right (position 14).

Obviously the opposite also happens, and the ORF is shortened. For example, consider gene (3.5) and suppose a mutation occurred at position 5, changing the "*" into "a":

```
0123456789012345 67890
+Q-/baaaQbaabaabbaaab           (3.8)
```

Its expression results in the following ET:

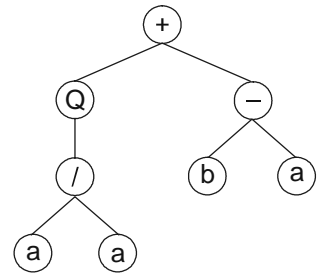

In this case, the ORF ends at position 7, shortening the original ET by 3 nodes.

Despite its fixed length, each gene has the potential to code for ETs of different sizes and shapes, the simplest being composed of only one node (when the first element of a gene is a terminal) and the biggest composed of as many nodes as the length of the gene (when all the elements of the head are functions with the maximum number of arguments, $n$).

It is evident from the examples above, that any modification made in the genome, no matter how profound, always results in a valid ET. Obviously the structural organization of genes must be preserved, always maintaining the boundaries between head and tail and not allowing symbols from the function set on the tail. Section 5 shows how GEP operators work and how they modify the genome of GEP individuals during reproduction.

## 3.3. Multigenic chromosomes

GEP chromosomes are usually composed of more than one gene of equal length. For each problem or run, the number of genes, as well as the length of the head, is chosen. Each gene codes for a sub-ET and the sub-ETs interact with one another forming a more complex multisubunit ET. The details of such interactions are fully explained in section 3.4.

Consider, for example, the following chromosome with length 27, composed of three genes (the tails are shown in bold):

```
012345678012345678012345678
-b*babbab*Qb+abbba-*Qabbaba          (3.9)
```

It has three ORFs, and each ORF codes for a sub-ET (Figure 2). Position 0 marks the start of each gene; the end of each ORF, though, is only evident upon construction of the respective sub-ET. As shown in Figure 2, the first ORF ends at position 4 (sub-$ET_1$); the second ORF ends at position 5 (sub-$ET_2$); and the last ORF also ends at position 5 (sub-$ET_3$). Thus, GEP chromosomes code for one or more ORFs, each expressing a particular sub-ET. Depending on the task at hand, these sub-ETs may be selected individually according to their respective fitness (e.g., in problems with multiple outputs), or they may form a more complex, multi-subunit ET and be selected according to the fitness of the whole, multi-subunit ET. The patterns of expression and the details of selection will be discussed throughout this paper. However, keep in mind that each sub-ET is both a separate entity and a part of a more complex, hierarchical structure, and, as in all complex systems, the whole is more than the sum of its parts.

## 3.4. Expression trees and the phenotype

In nature, the phenotype has multiple levels of complexity, the most complex being the organism itself. But tRNAs, proteins, ribosomes, cells, and so forth, are also products of expression, and all of them are ultimately encoded in the genome. In all cases, however, the expression of the genetic information starts with transcription (the synthesis of RNA) and, for protein genes, proceeds with translation (the synthesis of proteins).

### 3.4.1. Information decoding: Translation

In GEP, from the simplest individual to the most complex, the expression of genetic information starts with translation, the transfer of information from a gene into an ET. This process has already been presented in section 3.2 where decoding of GEP genes is shown. In contrast to nature, the expression of the genetic information in GEP is very simple. Worth emphasizing is the fact that in GEP there is no need for transcription: the message in the gene is directly translated into an ET.

GEP chromosomes are composed of one or more ORFs, and obviously the encoded individuals have different degrees of complexity. The simplest individuals are encoded in a single gene, and the "organism" is, in this case, the product of a single gene - an ET. In other cases, the organism is a multi-subunit ET, in which the different sub-ETs are linked together by a particular function. In other cases, the organism emerges from the spatial organization of different sub-ETs (e.g., in planning and problems with multiple outputs). And, in yet other cases, the organism emerges from the interactions of conventional sub-ETs with different domains (e.g., neural networks). However, in all cases, the whole organism is encoded in a linear genome.

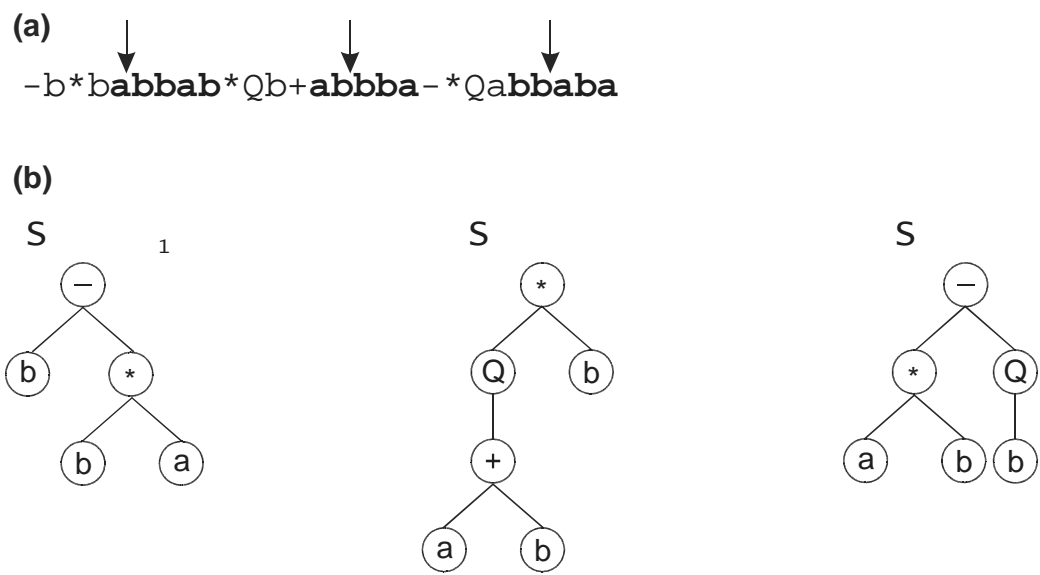

**Figure 2.** Expression of GEP genes as sub-ETs. **(a)** A three-genic chromosome with the tails shown in bold. The arrows show the termination point of each gene. **(b)** The sub-ETs codified by each gene.

*3.4.2. Interactions of sub-expression trees*

We have seen that translation results in the formation of sub-ETs with different complexity, but the complete expression of the genetic information requires the interaction of these sub-ETs with one another. One of the simplest interactions is the linking of sub-ETs by a particular function. This process is similar to the assemblage of different protein subunits into a multi-subunit protein.

When the sub-ETs are algebraic or boolean expressions, any mathematical or boolean function with more than one argument can be used to link the sub-ETs into a final, multi-subunit ET. The functions most chosen are addition or multiplication for algebraic sub-ETs, and OR or IF for boolean sub-ETs.

In the current version of GEP the linking function is *a priori* chosen for each problem, but it can be easily introduced in the genome; for instance, in the last position of chromosomes, and also be subjected to adaptation. Indeed, preliminary results suggest that this system works very well.

Figure 3 illustrates the linking of two sub-ETs by addition. Note that the root of the final ET (+) is not encoded by the genome. Note also that the final ET could be linearly encoded as the following K-expression:

```
0123456789012
+Q**-bQ+abbba
```
(3.10)

However, to evolve solutions for complex problems, it is more effective to use multigenic chromosomes, for they permit the modular construction of complex, hierarchical structures, where each gene codes for a small building block.

These small building blocks are separated from each other, and thus can evolve independently. For instance, if we tried to evolve a solution for the symbolic regression problem presented in section 6.1 with single-gene chromosomes, the success rate would fall significantly (see section 6.1). In that case the discovery of small building blocks is more constrained as they are no longer free to evolve independently. This kind of experiment shows that GEP is in effect a powerful, hierarchical invention system capable of easily evolving simple blocks and using them to form more complex structures [8, 9].

Figure 4 shows another example of sub-ET interaction, where three boolean sub-ETs are linked by the function IF. The multi-subunit ET could be linearized as the following K-expression:

```
01234567890123456789012
IINAIAINu1ca3aa2acAOab2
```
(3.11)

Figure 5 shows another example of sub-ET interaction, where the sub-ETs are of the simplest kind (one-element sub-ETs). In this case, the sub-ETs are linked 3 by 3 with the IF function, then these clusters are, in their turn, linked also 3 by 3 with another IF function, and the three last clusters are also linked by IF, forming a large multi-subunit ET. This kind of chromosomal architecture was used to evolve solutions for the 11-multiplexer problem of section 6.5.2 and also to evolve cellular automata rules for the density-classification problem. The individual of Figure 5 could be converted into the following K-expression:

```
IIIIIIIIIIIII131u3ab2ubab23c3ua31a333au3
```
(3.12)

And finally, the full expression of certain chromosomes requires the sequential execution of small plans, where the

**(a)**
```
012345678012345678
Q*Q+bbaaa*-babaabb
```

**(b)** 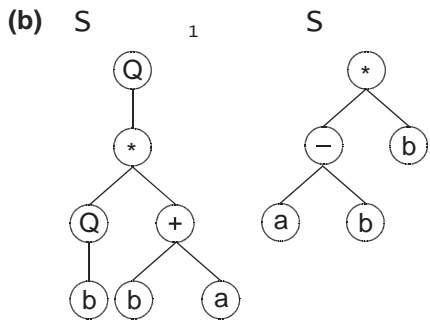

**(c)** 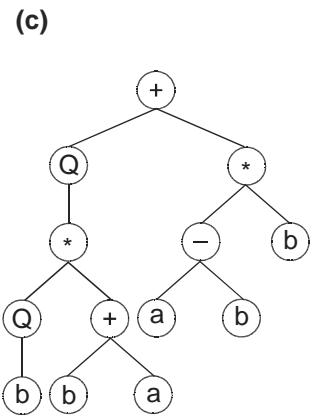

**Figure 3.** Expression of multigenic chromosomes as ETs. **(a)** A two-genic chromosome with the tails shown in bold. **(b)** The sub-ETs codified by each gene. **(c)** The result of posttranslational linking with addition.

**(a)**

```
IIAIca3aa2acuNNAOab2u3c31cAu12ua3112cac
```

**(b)** Sub-ET₁     Sub-ET     Sub-ET

**(c)** ET

**Figure 4.** Expression of multigenic chromosomes as ETs.
**(a)** A three-genic chromosome with the tails shown in bold ("N" is a function of one argument and represents NOT; "A" and "O" are functions of two arguments and represent respectively AND and OR; "I" is a function of three arguments and represents IF; the remaining symbols are terminals). **(b)** The sub-ETs codified by each gene. **(c)** The result of posttranslational linking with IF.

**(a)**

```
131u3ab2ubab23c3ua31a333au3
```

**(b)** ET

**Figure 5.** Expression of multigenic chromosomes as ETs. **(a)** A 27-genic chromosome composed of one-element genes. **(b)** The result of posttranslational linking with IF.

first sub-ET does a little work, the second continues from that, and so on. The final plan results from the orderly action of all subplans (see the block stacking problem in section 6.3).

The type of linking function, as well as the number of genes and the length of each gene, are *a priori* chosen for each problem. So, we can always start by using a single-gene chromosome, gradually increasing the length of the head; if it becomes very large, we can increase the number of genes and of course choose a function to link them. We can start with addition or OR, but in other cases another linking function might be more appropriate. The idea, of course, is to find a good solution, and GEP provides the means of finding one.

## 4. Fitness functions and selection

In this section, two examples of fitness functions are described. Other examples of fitness functions are given in the problems studied in section 6. The success of a problem greatly depends on the way the fitness function is designed: the goal must be clearly and correctly defined in order to make the system evolve in that direction.

### 4.1. Fitness functions

One important application of GEP is symbolic regression or function finding (e.g., [9]), where the goal is to find an expression that performs well for all fitness cases within a certain error of the correct value. For some mathematical applications it is useful to use small relative or absolute errors in order to discover a very good solution. But if the range of selection is excessively narrowed, populations evolve very slowly and are incapable of finding a correct solution. On the other hand, if the opposite is done and the range of selection is broadened, numerous solutions will appear with maximum fitness that are far from good solutions.

To solve this problem, an evolutionary strategy was devised that permits the discovery of very good solutions without halting evolution. So, the system is left to find for itself the best possible solution within a minimum error. For that a very broad limit for selection to operate is given, for instance, a relative error of 20%, that allows the evolutionary process to get started. Indeed, these founder individuals are usually very unfit but their modified descendants are reshaped by selection and populations adapt wonderfully, finding better solutions that progressively approach a perfect solution. Mathematically, the fitness $f_i$ of an individual program $i$ is expressed by equation (4.1a) if the error chosen is the absolute error, and by equation (4.1b) if the error chosen is the relative error:

$$f_i = \sum_{j=1}^{C_t} \left( M - \left| C_{(i,j)} - T_{(j)} \right| \right) \quad (4.1a)$$

$$f_i = \sum_{j=1}^{C_t} \left( M - \left| \frac{C_{(i,j)} - T_{(j)}}{T_{(j)}} \cdot 100 \right| \right) \quad (4.1b)$$

where $M$ is the range of selection, $C_{(i,j)}$ the value returned by the individual chromosome $i$ for fitness case $j$ (out of $C_t$ fitness cases), and $T_j$ is the target value for fitness case $j$. Note that for a perfect fit $C_{(i,j)} = T_j$ and $f_i = f_{\max} = C_t \cdot M$. Note that with this kind of fitness function the system can find the optimal solution for itself.

In another important GEP application, boolean concept learning or logic synthesis (e.g., [9]), the fitness of an individual is a function of the number of fitness cases on which it performs correctly. For most boolean applications, though, it is fundamental to penalize individuals able to solve correctly about 50% of fitness cases, as most probably this only reflects the 50% likelihood of correctly solving a binary boolean function. So, it is advisable to select only individuals capable of solving more than 50 to 75% of fitness cases. Below that mark a symbolic value of fitness can be attributed, for instance $f_i = 1$. Usually, the process of evolution is put in motion with these unfit individuals, for they are very easily created in the initial population. However, in future generations, highly fit individuals start to appear, rapidly spreading in the population. For easy problems, like boolean functions with 2 through 5 arguments, this is not really important, but for more complex problems it is convenient to choose a bottom line for selection. For these problems, the following fitness function can be used:

$$\text{If } n \geq \tfrac{1}{2} C_t, \text{ then } f_t = n; \text{ else } f_t = 1 \quad (4.2)$$

where $n$ is the number of fitness cases correctly evaluated, and $C_t$ is the total number of fitness cases.

### 4.2. Selection

In all the problems presented in this work, individuals were selected according to fitness by roulettewheel sampling [10] coupled with the cloning of the best individual (simple elitism). A preliminary study of different selection schemes (roulettewheel selection with and without elitism, tournament selection with and without elitism, and various kinds of deterministic selection with and without elitism) suggests that there is no appreciable difference between them as long as the cloning of the best individual is guaranteed (results not shown). Some schemes perform better in one problem, others in another. However, for more complex problems it seems that roulettewheel selection with elitism is best.

## 5. Reproduction with modification

According to fitness and the luck of the roulette, individuals are selected to reproduce with modification, creating the necessary genetic diversification that allows evolution in the long run.

Except for replication, where the genomes of all the selected individuals are rigorously copied, all the remaining operators randomly pick chromosomes to be subjected to a certain modification. However, except for mutation, each operator is not allowed to modify a chromosome more than once. For instance, for a transposition rate of 0.7, seven out of 10 different chromosomes are randomly chosen.

Furthermore, in GEP, a chromosome might be chosen by none or several genetic operators that introduce variation in the population. This feature also distinguishes GEP from GP where an entity is never modified by more than one operator at a time [9]. Thus, in GEP, the modifications of several genetic operators accumulate during reproduction, producing offspring very different from the parents.

We now proceed with the detailed description of GEP operators, starting obviously with replication. (Readers less concerned with implementation details of genetic operators may wish to skip this section.)

### 5.1. Replication

Although vital, replication is the most uninteresting operator: alone it contributes nothing to genetic diversification. (Indeed, replication, together with selection, is only capable of causing genetic drift.) According to fitness and the luck of the roulette, chromosomes are faithfully copied into the next generation. The fitter the individual the higher the probability of leaving more offspring. Thus, during replication the genomes of the selected individuals are copied as many times as the outcome of the roulette. The roulette is spun as many times as there are individuals in the population, always maintaining the same population size.

### 5.2. Mutation

Mutations can occur anywhere in the chromosome. However, the structural organization of chromosomes must remain intact. In the heads any symbol can change into another (function or terminal); in the tails terminals can only change into terminals. This way, the structural organization of chromosomes is maintained, and all the new individuals produced by mutation are structurally correct programs.

Typically, a mutation rate ($p_m$) equivalent to two point mutations per chromosome is used. Consider the following 3-genic chromosome:

```
012345678012345678012345678
-+-+abaaa/bb/ababb*Q*+aaaba
```

Suppose a mutation changed the element in position 0 in gene 1 to "Q"; the element in position 3 in gene 2 to "Q"; and the element in position 1 in gene 3 to "b", obtaining:

```
012345678012345678012345678
Q+-+abaaa/bbQababb*b*+aaaba
```

Note that if a function is mutated into a terminal or *vice versa*, or a function of one argument is mutated into a function of two arguments or *vice versa*, the ET is modified drastically. Note also that the mutation on gene 2 is an example of a neutral mutation, as it occurred in the noncoding region of the gene.

It is worth noticing that in GEP there are no constraints neither in the kind of mutation nor the number of mutations in a chromosome: in all cases the newly created individuals are syntactically correct programs.

In nature, a point mutation in the sequence of a gene can slightly change the structure of the protein or not change it at all, as neutral mutations are fairly frequent (e.g., mutations in introns, mutations that result in the same amino acid due to the redundancy of the genetic code, etc.). Here, although neutral mutations exist (e.g., mutations in the noncoding regions), a mutation in the coding sequence of a gene has a much more profound effect: it usually drastically reshapes the ET.

### 5.3. Transposition and insertion sequence elements

The transposable elements of GEP are fragments of the genome that can be activated and jump to another place in the chromosome. In GEP there are three kinds of transposable elements. (1) Short fragments with a function or terminal in the first position that transpose to the head of genes, except to the root (insertion sequence elements or IS elements). (2) Short fragments with a function in the first position that transpose to the root of genes (root IS elements or RIS elements). (3) Entire genes that transpose to the beginning of chromosomes.

The existence of IS and RIS elements is a remnant of the developmental process of GEP, as the first GEA used only single-gene chromosomes, and in such systems a gene with a terminal at the root was of little use. When multigenic chromosomes were introduced this feature remained as these operators are important to understand the mechanisms of genetic variation and evolvability.

#### 5.3.1. Transposition of insertion sequence elements

Any sequence in the genome might become an IS element, therefore these elements are randomly selected throughout the chromosome. A copy of the transposon is made and inserted at any position in the head of a gene, except at the start position.

Typically, an IS transposition rate ($p_{is}$) of 0.1 and a set of three IS elements of different length are used. The transposition operator randomly chooses the chromosome, the start of the IS element, the target site, and the length of the transposon. Consider the 2-genic chromosome below:

```
012345678901234567890012345678901234567890
*-+*a-+a*bbabbaabababQ**+abQbb*aabbaaaabba
```

Suppose that the sequence "bba" in gene 2 (positions 12 through 14) was chosen to be an IS element, and the target site was bond 6 in gene 1 (between positions 5 and 6). Then, a cut is made in bond 6 and the block "bba" is copied into the site of insertion, obtaining:

```
0123456789012345678900123456789012345678 90
*-+*a-bba+babbaabababQ**+abQbb*aabbaaaabba
```

During transposition, the sequence upstream from the insertion site stays unchanged, whereas the sequence downstream from the copied IS element loses, at the end of the head, as many symbols as the length of the IS element (in this case the sequence "a*b" was deleted). Note that, despite this insertion, the structural organization of chromosomes is maintained, and therefore all newly created individuals are syntactically correct programs. Note also that transposition can drastically reshape the ET, and the more upstream the insertion site the more profound the change. Thus, this kind of operator (IS transposition and RIS transposition below) may be seen as having a high hit rate at the lowest levels of ETs [7].

*5.3.2. Root transposition*

All RIS elements start with a function, and thus are chosen among the sequences of the heads. For that, a point is randomly chosen in the head and the gene is scanned downstream until a function is found. This function becomes the start position of the RIS element. If no functions are found, it does nothing.

Typically a root transposition rate ($p_{ris}$) of 0.1 and a set of three RIS elements of different sizes are used. This operator randomly chooses the chromosome, the gene to be modified, the start of the RIS element, and its length. Consider the following 2-genic chromosome:

```
0123456789012345678900123456789012345678 90
-ba*+-+-Q/ababababbbaaaQ*b/+bbabbaaaaaaaabbb
```

Suppose that the sequence "+bb" in gene 2 was chosen to be an RIS element. Then, a copy of the transposon is made into the root of the gene, obtaining:

```
0123456789012345678900123456789012345678 90
-ba*+-+-Q/ababababbbaaa+bbQ*b/+bbaaaaaaaabbb
```

During root transposition, the whole head shifts to accommodate the RIS element, losing, at the same time, the last symbols of the head (as many as the transposon length). As with IS elements, the tail of the gene subjected to transposition and all nearby genes stay unchanged. Note, again, that the newly created programs are syntactically correct because the structural organization of the chromosome is maintained.

The modifications caused by root transposition are extremely radical, because the root itself is modified. In nature, if a transposable element is inserted at the beginning of the coding sequence of a gene, causing a frameshift mutation, it radically changes the encoded protein. Like mutation and IS transposition, root insertion has a tremendous transforming power and is excellent for creating genetic variation.

*5.3.3. Gene transposition*

In gene transposition an entire gene functions as a transposon and transposes itself to the beginning of the chromosome. In contrast to the other forms of transposition, in gene transposition the transposon (the gene) is deleted in the place of origin. This way, the length of the chromosome is maintained.

The chromosome to undergo gene transposition is randomly chosen, and one of its genes (except the first, obviously) is randomly chosen to transpose. Consider the following chromosome composed of 3 genes:

```
012345678012345678012345678
*a-*abbab-QQ/aaabbQ+abababb
```

Suppose gene 2 was chosen to undergo gene transposition. Then the following chromosome is obtained:

```
012345678012345678012345678
-QQ/aaabb*a-*abbabQ+abababb
```

Note that for numerical applications where the function chosen to link the genes is addition, the expression evaluated by the chromosome is not modified. But the situation differs in other applications where the linking function is not commutative, for instance, the IF function chosen to link the sub-ETs in the 11-multiplexer problem in section 6.5.2. However, the transforming power of gene transposition reveals itself when this operator is conjugated with crossover. For example, if two functionally identical chromosomes or two chromosomes with an identical gene in different positions recombine, a new individual with a duplicated gene might appear. It is known that the duplication of genes plays an important role in biology and evolution (e.g., [11]). Interestingly, in GEP, individuals with duplicated genes are commonly found in the process of problem solving.

*5.4. Recombination*

In GEP there are three kinds of recombination: one-point, two-point, and gene recombination. In all cases, two parent chromosomes are randomly chosen and paired to exchange some material between them.

*5.4.1. One-point recombination*

During one-point recombination, the chromosomes cross over a randomly chosen point to form two daughter chromosomes. Consider the following parent chromosomes:

```
012345678012345678
-b+Qbbabb/aQbbbaab
/-a/ababb-ba-abaaa
```

Suppose bond 3 in gene 1 (between positions 2 and 3) was randomly chosen as the crossover point. Then, the paired chromosomes are cut at this bond, and exchange between them the material downstream from the crossover point, forming the offspring below:

```
012345678012345678
-b+/ababb-ba-abaaa
/-aQbbabb/aQbbbaab
```

With this kind of recombination, most of the time, the offspring created exhibit different properties from those of the parents. One-point recombination, like the above mentioned operators, is a very important source of genetic variation, being, after mutation, one of the operators most chosen in GEP. The one-point recombination rate ($p_{1r}$) used depends on the rates of other operators. Typically a global crossover rate of 0.7 (the sum of the rates of the three kinds of recombination) is used.

*5.4.2. Two-point recombination*

In two-point recombination the chromosomes are paired and the two points of recombination are randomly chosen. The material between the recombination points is afterwards exchanged between the two chromosomes, forming two new daughter chromosomes. Consider the following parent chromosomes:

```
012345678901234567890
+*a*bbcccac*baQ*acabab-[1]
*cbb+cccbcc++**bacbaab-[2]
```

Suppose bond 7 in gene 1 (between positions 6 and 7) and bond 3 in gene 2 (between positions 2 and 3) were chosen as the crossover points. Then, the paired chromosomes are cut at these bonds, and exchange the material between the crossover points, forming the offspring below:

```
012345678901234567890
+*a*bbccbcc++*Q*acabab-[3]
*cbb+ccccac*ba*bacbaab-[4]
```

Note that the first gene is, in both parents, split downstream from the termination point. Indeed, the noncoding regions of GEP chromosomes are ideal regions where chromosomes can be split to cross over without interfering with the ORFs. Note also that the second gene of chromosome 1 was also cut downstream from the termination point. However, gene 2 of chromosome 2 was split upstream from the termination point, profoundly changing the sub-ET. Note also that when these chromosomes recombined, the noncoding region of gene 2 of chromosome 1 was activated and integrated into chromosome 3.

The transforming power of two-point recombination is greater than one-point recombination, and is most useful to evolve solutions for more complex problems, especially when multigenic chromosomes composed of several genes are used.

*5.4.3. Gene recombination*

In gene recombination an entire gene is exchanged during crossover. The exchanged genes are randomly chosen and occupy the same position in the parent chromosomes. Consider the following parent chromosomes:

```
012345678012345678012345678
/aa-abaaa/a*bbaaab/Q*+aaaab
/-*/abbabQ+aQbabaa-Q/Qbaaba
```

Suppose gene 2 was chosen to be exchanged. In this case the following offspring is formed:

```
012345678012345678012345678
/aa-abaaaQ+aQbabaa/Q*+aaaab
/-*/abbab/a*bbaaab-Q/Qbaaba
```

The newly created individuals contain genes from both parents. Note that with this kind of recombination, similar genes can be exchanged but, most of the time, the exchanged genes are very different and new material is introduced into the population.

It is worth noting that this operator is unable to create new genes: the individuals created are different arrangements of existing genes. In fact, when gene recombination is used as the unique source of genetic variation, more complex problems can only be solved using very large initial populations in order to provide for the necessary diversity of genes (see section 6.1). However, the creative power of GEP is based not only in the shuffling of genes or building blocks, but also in the constant creation of new genetic material.

## 6. Six examples of gene expression programming in problem solving

The suite of problems chosen to illustrate the functioning of this new algorithm is quite varied, including not only problems from different fields (symbolic regression, planning, Boolean concept learning, and cellular automata rules) but also problems of great complexity (cellular automata rules for the density-classification task).

*6.1. Symbolic regression*

The objective of this problem is the discovery of a symbolic expression that satisfies a set of fitness cases. Consider we are given a sampling of the numerical values from the function

$$y = a^4 + a^3 + a^2 + a \tag{6.1}$$

over 10 chosen points and we want to find a function fitting those values within 0.01 of the correct value.

First, the set of functions $F$ and the set of terminals $T$ must be chosen. In this case $F = \{+, -, *, /\}$ and $T = \{a\}$. Then the structural organization of chromosomes, namely the length of the head and the number of genes, is chosen. It is advisable to start with short, single-gene chromosomes and then gradually increase $h$. Figure 6 shows such an analysis for this problem. A population size $P$ of 30 individuals and an evolutionary time $G$ of 50 generations were used. A $p_m$ equivalent to two one-point mutations per chromosome and a $p_{1r} = 0.7$ were used in all the experiments in order to simplify the analysis. The set of fitness cases is shown in Table 1 and the fitness was evaluated by equation (4.1a), being $M = 100$. If $|C_{(i,j)} - T_j|$ is equal to or less than 0.01 (the precision), then $|C_{(i,j)} - T_j| = 0$ and $f_{(i,j)} = 100$; thus for $C_t = 10$, $f_{max} = 1000$.

Note that GEP can be useful in searching the most parsimonious solution to a problem. For instance, the chromosome

```
0123456789012
*++/**aaaaaaa
```

with $h = 6$ codes for the ET:

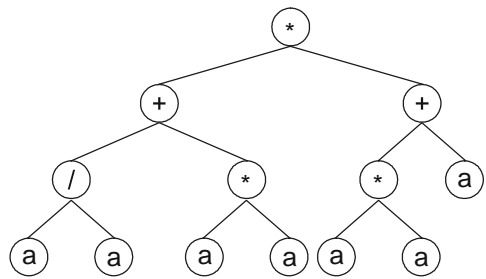

which is equivalent to the target function. Note also that GEP can efficiently evolve solutions using large values of $h$, that is, it is capable of evolving large and complex sub-ETs. It is worth noting that the most compact genomes are not the most efficient. Therefore a certain redundancy is fundamental to efficiently evolve good programs.

In another analysis, the relationship between success rate and population size $P$, using an $h = 24$ was studied (Figure 7). These results show the supremacy of a genotype/pheno-type representation, as this single-gene system, which is equivalent to GP, greatly surpasses that technique [9]. However, GEP is much more complex than a single-gene system because GEP chromosomes can encode more than one gene (see Figure 8).

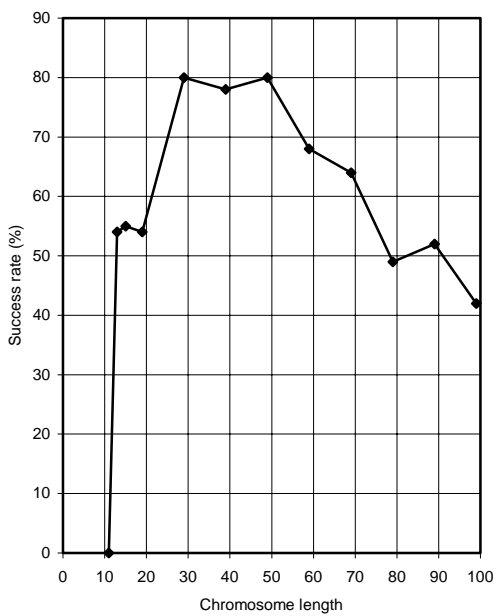

**Figure 6.** Variation of success rate ($P_s$) with chromosome length. For this analysis $G = 50$, $P = 30$, and $P_s$ was evaluated over 100 identical runs.

**Table 1**
Set of fitness cases for the symbolic regression problem.

| a | f(a) |
|---|---|
| 2.81 | 95.2425 |
| 6 | 1554 |
| 7.043 | 2866.55 |
| 8 | 4680 |
| 10 | 11110 |
| 11.38 | 18386 |
| 12 | 22620 |
| 14 | 41370 |
| 15 | 54240 |
| 20 | 168420 |

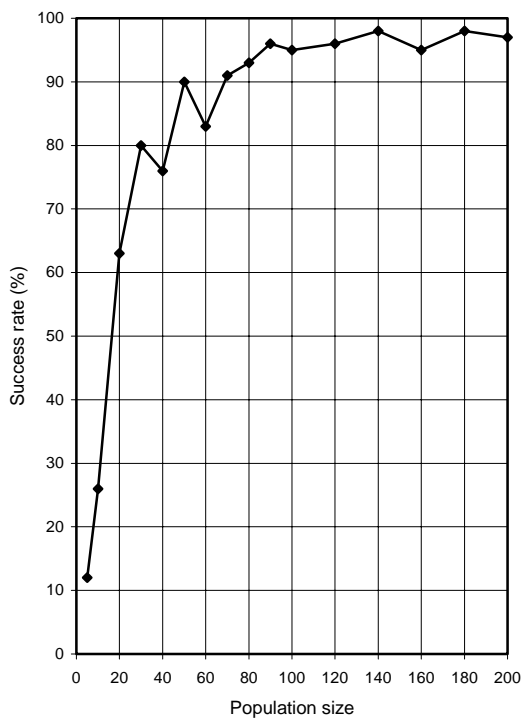

**Figure 7.** Variation of success rate ($P_s$) with population size. For this analysis $G = 50$, and a medium value of 49 for chromosome length ($h = 24$) was used. $P_s$ was evaluated over 100 identical runs.

Suppose we could not find a solution after the analysis shown in Figure 6. Then we could increase the number of genes, and choose a function to link them. For instance, we could choose an $h = 6$ and then increase the number of genes gradually. Figure 8 shows how the success rate for this problem depends on the number of genes. In this analysis, the $p_m$ was equivalent to two one-point mutations per chromosome, $p_{1r} = 0.2$, $p_{2r} = 0.5$, $p_{gr} = 0.1$, $p_{is} = 0.1$, $p_{ris} = 0.1$, $p_{gt} = 0.1$, and three transposons (both IS and RIS elements) of lengths 1, 2, and 3 were used. Note that GEP can cope very well with an excess of genes: the success rate for the 10-genic system is still very high (47%).

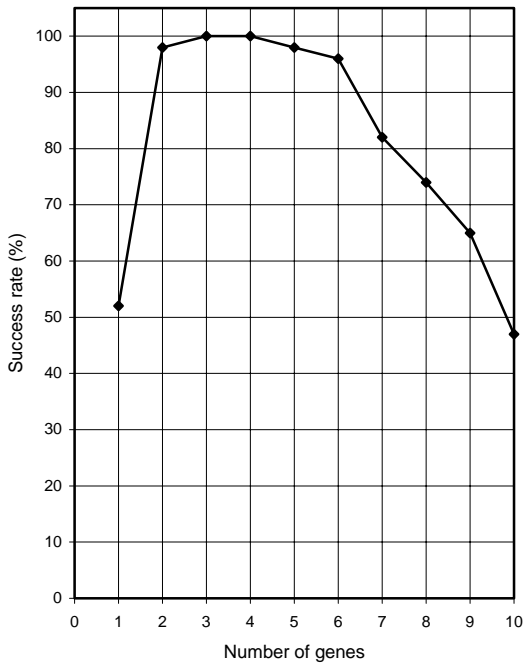

**Figure 8.** Variation of success rate ($P_s$) with the number of genes. For this analysis $G = 50$, $P = 30$, and $h = 6$ (a gene length of 13). $P_s$ was evaluated over 100 identical runs.

In Figure 9 another important relationship is shown: how the success rate depends on evolutionary time. In contrast to GP where 51 generations are the norm, for after that nothing much can possibly be discovered [7], in GEP, populations can adapt and evolve indefinitely because new material is constantly being introduced into the genetic pool.

Finally, suppose that the multigenic system with sub-ETs linked by addition could not evolve a satisfactory solution. Then we could choose another linking function, for instance, multiplication. This process is repeated until a good solution has been found.

As stated previously, GEP chromosomes can be easily modified in order to encode the linking function as well. In this case, for each problem the ideal linking function would be found in the process of adaptation.

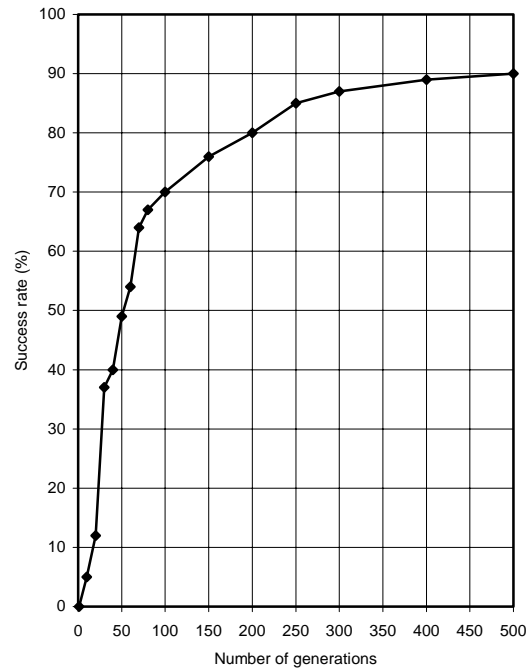

**Figure 9.** Variation of success rate ($P_s$) with the number of generations. For this analysis $P = 30$, $p_m = 0.051$, $p_{1r} = 0.7$ and a chromosome length of 79 (a single-gene chromosome with $h = 39$) was used. $P_s$ was evaluated over 100 identical runs.

Consider, for instance, a multigenic system composed of 3 genes linked by addition. As shown in Figure 8, the success rate has in this case the maximum value of 100%. Figure 10 shows the progression of average fitness of the population and the fitness of the best individual for run 0 of the experiment summarized in Table 2, column 1. In this run, a correct solution was found in generation 11. The sub-ETs are linked by addition:

```
0123456789012012345678901201234566789012
**-*a+aaaaaaa++**a*aaaaaaa*+-a/aaaaaaa
```

and mathematically corresponds to the target function (the contribution of each sub-ET is indicated in brackets):

$$y = (a^4) + (a^3 + a^2 + a) + (0) = a^4 + a^3 + a^2 + a.$$

The detailed analysis of this program shows that some of the actions are redundant for the problem at hand, like the addition of 0 or multiplication by 1. However, the existence of these unnecessary clusters, or even pseudogenes like gene 3, is important to the evolution of more fit individuals (compare, in Figures 6 and 8, the success rate of a compact, single-gene system with $h = 6$ with other less compact systems both with more genes and $h$ greater than 6).

The plot for average fitness in Figure 10 (and also Figures 12, 13 and 17 below) suggests different evolutionary dynamics for GEP populations. The oscillations on average fitness, even after the discovery of a perfect solution, are unique to GEP. A certain degree of oscillation is due to the small population sizes used to solve the problems presented in this work. However, an identical pattern is obtained using larger population sizes. Figure 11 compares six evolutionary dynamics in populations of 500 individuals for 500 generations. Plot 1 (all operators active) shows the progression of average fitness of an experiment identical to the one summarized in Table 2, column 1, that is, with all the genetic operators switched on. The remaining dynamics were obtained for mutation alone (Plot 2), for gene recombination combined with gene transposition (Plot 3), for one-point recombination (Plot 4), two-point recombination (Plot 5), and gene recombination (Plot 6).

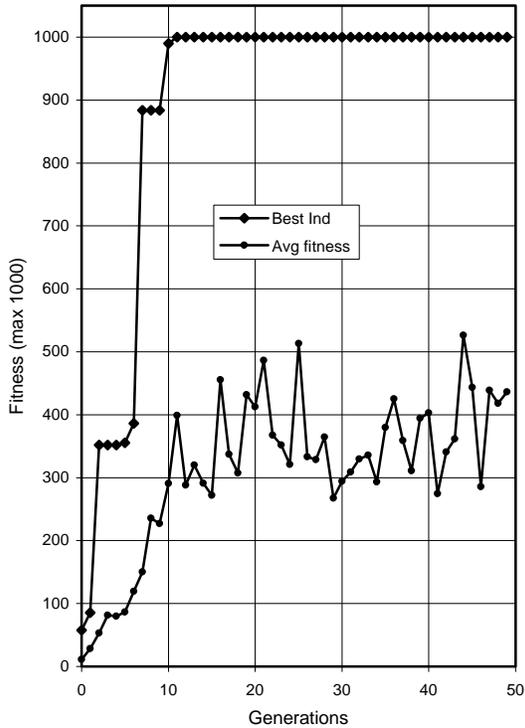

**Figure 10.** Progression of average fitness of the population and the fitness of the best individual for run 0 of the experiment summarized in Table 2, column 1 (symbolic regression).

It is worth noticing the homogenizing effect of all kinds of recombination. Interestingly, this kind of pattern is similar to the evolutionary dynamics of GAs and GP populations [9, 10]. Also worth noticing is the plot for gene recombination alone (Figure 11, Plot 6): in this case a perfect solution was not found. This shows that sometimes it is impossible to find a perfect solution only by shuffling existing building blocks, as is done in all GP implementations without mutation. Indeed, GEP gene recombination is similar in effect to GP recombination, for it permits exclusively the recombination of mathematically concise blocks. Note that even a more generalized shuffling of building blocks (using gene recombination combined with gene transposition) results in oscillatory dynamics (Figure 11, Plot 3).

**Table 2**
Parameters for the symbolic regression (SR), sequence induction (SI), sequence induction using ephemeral random constants (SI*), block stacking (BS), and 11-multiplexer (11-M) problems.

|  | SR | SI | SI* | BS | 11-M |
|---|---|---|---|---|---|
| Number of runs | 100 | 100 | 100 | 100 | 100 |
| Number of generations | 50 | 100 | 100 | 100 | 400 |
| Population size | 30 | 50 | 50 | 30 | 250 |
| Number of fitness cases | 10 | 10 | 10 | 10 | 160 |
| Head length | 6 | 6 | 7 | 4 | 1 |
| Number of genes | 3 | 7 | 8 | 3 | 27 |
| Chromosome length | 39 | 91 | 184 | 27 | 27 |
| Mutation rate | 0.051 | 0.022 | 0.011 | 0.074 | 0.074 |
| One-point recombination rate | 0.2 | 0.7 | 0.5 | 0.1 | 0.7 |
| Two-point recombination rate | 0.5 | 0.1 | 0.2 | -- | -- |
| Gene recombination rate | 0.1 | 0.1 | 0.1 | 0.7 | -- |
| IS transposition rate | 0.1 | 0.1 | 0.1 | 0.1 | -- |
| IS elements length | 1,2,3 | 1,2,3 | 1 | 1 | -- |
| RIS transposition rate | 0.1 | 0.1 | 0.1 | 0.1 | -- |
| RIS elements length | 1,2,3 | 1,2,3 | 1 | 1 | -- |
| Gene transposition rate | 0.1 | 0.1 | 0.1 | -- | -- |
| Random constants mutation rate | -- | -- | 0.01 | -- | -- |
| Dc specific IS transposition rate | -- | -- | 0.013 | -- | -- |
| Selection range | 100 | 20% | 20% | -- | -- |
| Error | 0.01 | 0.0% | 0.0% | -- | -- |
| Success rate | 1 | 0.83 | 0.31 | 0.7 | 0.57 |

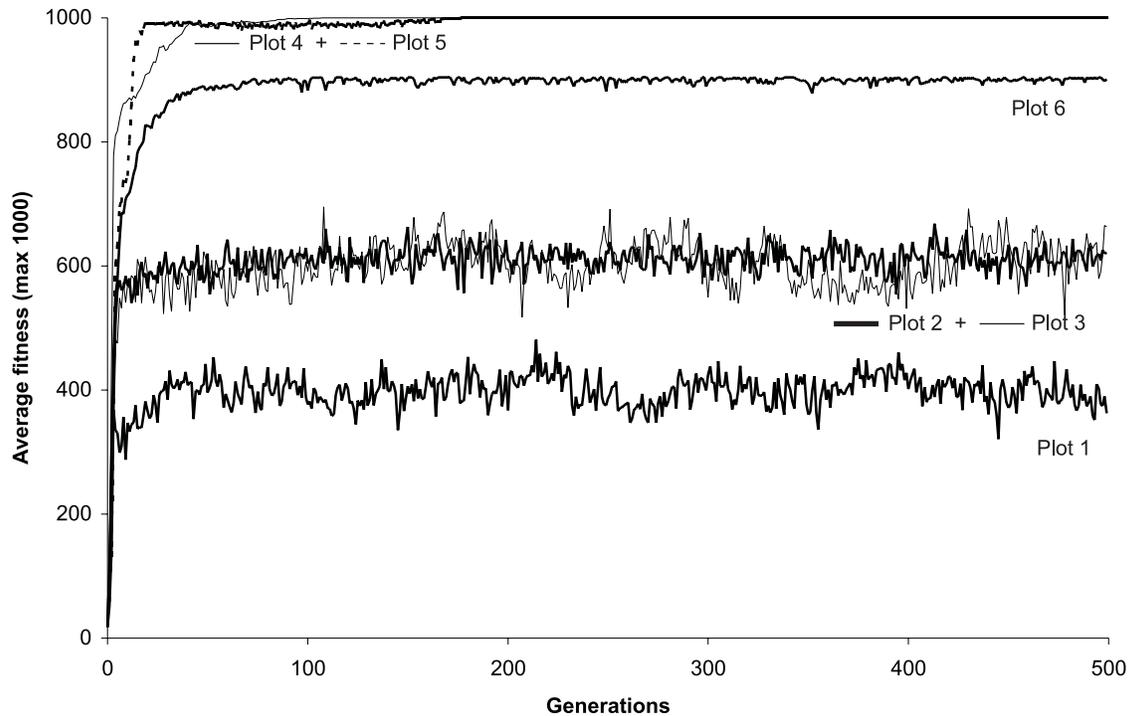

**Figure 11.** Possible evolutionary dynamics for GEP populations. For this analysis $P = 500$. The plots show the progression of average fitness of the population. Plot 1: All operators switched on with rates as shown in Table 2, column 1; in this case a perfect solution was found in generation 1. Plot 2: Only mutation at $p_m = 0.051$; in this case a perfect solution was found in generation 3. Plot 3: Only gene recombination at $p_{gr} = 0.7$ plus gene transposition at $p_{gt} = 0.2$ were switched on; in this case a perfect solution was found in generation 2. Plot 4: Only one-point recombination at $p_{1r} = 0.7$; in this case a perfect solution was found in generation 3. Plot 5: Only two-point recombination at $p_{2r} = 0.7$; in this case a perfect solution was found in generation 1. Plot 6: Only gene recombination at $p_{gr} = 0.7$; in this case a perfect solution was not found: the best of run has fitness 980 and was found in generation 2.

## 6.2. Sequence induction and the creation of constants

The problem of sequence induction is a special case of symbolic regression where the domain of the independent variable consists of the nonnegative integers. However, the sequence chosen is more complicated than the expression used in symbolic regression, as different coefficients were used.

The solution to this kind of problem involves the discovery of certain constants. Here two different approaches to the problem of constant creation are shown: one without using ephemeral random constants [9], and another using ephemeral random constants.

In the sequence 1, 15, 129, 547, 1593, 3711, 7465, 13539, 22737, 35983, 54321,..., the $n$th ($N$) term is

$$N = 5a_n^4 + 4a_n^3 + 3a_n^2 + 2a_n + 1 \qquad (6.2)$$

where $a_n$ consists of the nonnegative integers 0, 1, 2, 3,....

For this problem $F = \{+, -, *, /\}$ and $T = \{a\}$. The set of fitness cases is shown in Table 3 and the fitness was evaluated by equation (4.1b), being $M = 20\%$. Thus, if the 10 fitness cases were computed exactly, $f_{max} = 200$.

Figure 12 shows the progression of average fitness of the population and the fitness of the best individual for run 1 of the experiment summarized in Table 2, column 2. In this run, a perfect solution was found in generation 81 (the sub-ETs are linked by addition):

**Table 3**
Set of fitness cases for the sequence induction problem.

| a | N |
|---|---|
| 1 | 15 |
| 2 | 129 |
| 3 | 547 |
| 4 | 1593 |
| 5 | 3711 |
| 6 | 7465 |
| 7 | 13539 |
| 8 | 22737 |
| 9 | 35983 |
| 10 | 54321 |

```
0123456789012012345678901201234567890120123456789012...
*a/+a*aaaaaaa**-/**aaaaaaa**+++*aaaaaaa+-+a/*aaaaaaa...

           ...0123456789012012345678901201234567890120123456789012
           ...*a*-a+aaaaaaa-+++-+aaaaaaa+*/*/+aaaaaaa
```

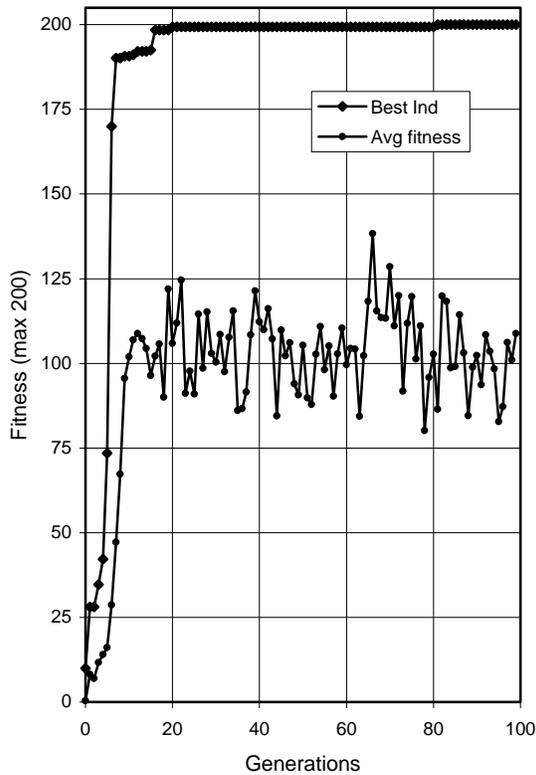

**Figure 12.** Progression of average fitness of the population and the fitness of the best individual for run 1 of the experiment summarized in Table 2, column 2 (sequence induction without ephemeral random constants).

and mathematically corresponds to the target sequence (the contribution of each sub-ET is indicated in brackets):

$$y = (a^2+a)+(a^4-a^3)+(4a^4+4a^3)+(a^2+2a-1)+(a^3)+(-a)+(a^2+2).$$

As shown in column 2 of Table 2, the probability of success for this problem using the first approach is 0.83. Note that all the constants are created from scratch by the algorithm. It seems that in real-world problems this kind of approach is more advantageous because, first, we never know beforehand what kind of constants are needed and, second, the number of elements in the terminal set is much smaller, reducing the complexity of the problem.

However, ephemeral random constants can be easily implemented in GEP. For that an additional domain Dc was created. Structurally, the Dc comes after the tail, has a length equal to $t$, and consists of the symbols used to represent the ephemeral random constants.

For each gene the constants are created at the beginning of a run, but their circulation is guaranteed by the genetic operators. Besides, a special mutation operator was created that allows the permanent introduction of variation in the set of random constants. A domain specific IS transposition was also created in order to guarantee an effective shuffling of the constants. Note that the basic genetic operators are not affected by the Dc: it is only necessary to keep the boundaries of each region and not mix different alphabets.

Consider the single-genic chromosome with an $h = 7$:

```
0123456789012345678 9012
*?**?+?aa??a?a?63852085          (6.3)
```

where "?" represents the ephemeral random constants. The expression of this kind of chromosome is done exactly as before, obtaining:

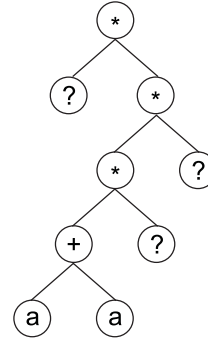

The "?" symbols in the ET are then replaced from left to right and from top to bottom by the symbols in Dc, obtaining:

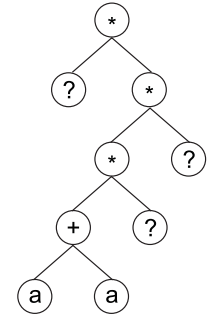

The values corresponding to these symbols are kept in an array. For simplicity, the number represented by the symbol indicates the order in the array. For instance, for the 10 element array

$$A = \{-0.004, 0.839, -0.503, 0.05, -0.49, -0.556, 0.43, \\ -0.899, 0.576, -0.256\}$$

the chromosome (6.3) above gives:

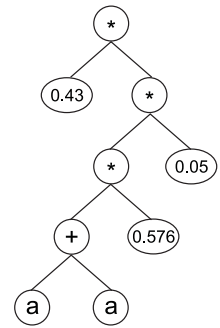

To solve the problem at hand using ephemeral random constants $F = \{+, -, *\}$, $T = \{a, ?\}$, the set of random constants $R = \{0, 1, 2, 3, 4, 5, 6, 7, 8, 9\}$, and the ephemeral random constant "?" ranged over the integers 0, 1, 2, and 3. The parameters used per run are shown in Table 2, column 3. In this experiment, the first solution was found in generation 91 of run 8 (the sub-ETs are linked by addition):

```
Gene 0: -??*a-*aaa?a?aa26696253
    A₀ = {3, 1, 0, 0, 3, 3, 2, 2, 2, 3}

Gene 1: *-aa-a-???a?aaa73834168
    A₁ = {0, 1, 2, 3, 1, 3, 0, 0, 1, 3}

Gene 2: +a??-+??aaaa?aa43960807
    A₂ = {1, 2, 1, 3, 3, 2, 2, 2, 1, 3}

Gene 3: *a***+aa?a??aaa20546809
    A₃ = {3, 0, 1, 3, 0, 2, 2, 2, 2, 0}

Gene 4: *a***+aa?aa?aaa34722724
    A₄ = {2, 3, 3, 2, 1, 3, 0, 0, 2, 3}

Gene 5: *a*++*+?aa??a?a54218512
    A₅ = {1, 3, 3, 1, 0, 0, 2, 0, 0, 2}

Gene 6: +a*?a*-a?aaa??a94759218
    A₆ = {3, 0, 0, 2, 1, 1, 3, 1, 3, 2}

Gene 7: +-?a*a??a?aa??a69085824
    A₇ = {2, 2, 3, 1, 3, 1, 0, 0, 1, 0}
```

and mathematically corresponds to the target function (the contribution of each sub-ET is indicated in brackets):

$$y = (-2)+(-3a)+(a+3)+(a^4+3a^3)+(4a^4)+(a^3+3a^2)+(3a).$$

As shown in column 3 of Table 2, the probability of success for this problem is 0.31, considerably lower than the 0.83 of the first approach. Furthermore, only the prior knowledge of the solution enabled us, in this case, to correctly choose the random constants. Therefore, for real-world applications where the magnitude and type of coefficients is unknown, it is more appropriate to let the system find the constants for itself. However, for some numerical applications the discovery of constants is fundamental and they can be easily created as indicated here.

### 6.3. Block stacking

In block stacking, the goal is to find a plan that takes any initial configuration of blocks randomly distributed between the stack and the table and places them in the stack in the correct order. In this case, the blocks are the letters of the word "universal". (Although the word universal was used as illustration, in this version the blocks being stacked may have identical labels like, for instance, in the word "individual".)

The functions and terminals used for this problem consisted of a set of actions and sensors, being $F = \{C, R, N, A\}$ (move to stack, remove from stack, not, and do until true, respectively), where the first three take one argument and "A" takes two arguments. In this version, the "A" loops are processed at the beginning and are solved in a particular order (from bottom to top and from left to right). The action argument is executed at least once despite the state of the predicate argument and each loop is executed only once, timing out after 20 iterations. The set of terminals consisted of three sensors {u, t, p} (current stack, top correct block, and next needed block, respectively). In this version, "t" refers only to the block on the top of the stack and whether it is correct or not; if the stack is empty or has some blocks, all of them correctly stacked, the sensor returns True, otherwise it returns False; and "p" refers obviously to the next needed block immediately after "t".

A multigenic system composed of three genes of length 9 was used in this problem. The linking of the sub-ETs consisted of the sequential execution of each sub-ET or sub-plan. For instance, if the first sub-ET empties all the stacks, the next sub-ET may proceed to fill them, and so on. The fitness was determined against 10 fitness cases (initial configurations of blocks). For each generation, an empty stack plus nine initial configurations with one to nine letters in the stack were randomly generated. The empty stack was used to prevent the untimely termination of runs, as a fitness point was attributed to each empty stack (see below). However, GEP is capable of efficiently solving this problem using 10 random initial configurations (results not shown).

The fitness function was as follows: for each empty stack one fitness point was attributed; for each partially and correctly packed stack (i.e., with 1 to 8 letters in the case of the word "universal") two fitness points were attributed; and for each completely and correctly stacked word 3 fitness points were attributed. Thus, the maximum fitness was 30. The idea was to make the population of programs hierarchically evolve solutions toward a perfect plan. And, in fact, usually the first useful plan discovered empties all the stacks, then some programs learn how to partially fill those empty stacks, and finally a perfect plan is discovered that fills the stacks completely and correctly (see Figure 13).

Figure 13 shows the progression of average fitness of the population and the fitness of the best individual for run 2 of the experiment summarized in Table 2, column 4. In this run, a perfect plan was found in generation 50:

```
012345678012345678012345678
ARCuptppuApNCptuutNtpRppptp
```

Note that the first sub-plan removes all the blocks and stacks a correct letter; the second sub-plan correctly stacks all the remaining letters; and the last sub-plan does nothing. It should be emphasized that the plans with maximum fitness evolved are in fact perfect, universal plans: each generation they are tested against nine randomly generated initial configurations, more than sufficient to allow the algorithm to generalize the problem (as shown in Figure 13, once reached, the maximum fitness is maintained). Indeed, with the fitness function and the kind of fitness cases used, all plans with maximum fitness are universal plans.

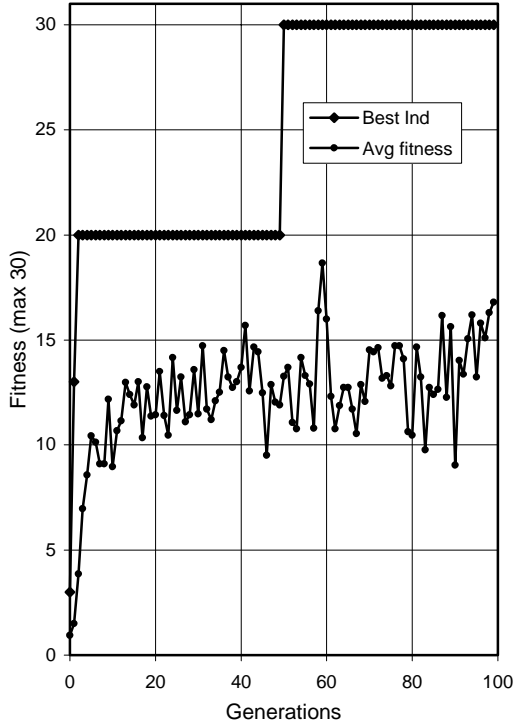

**Figure 13.** Progression of average fitness of the population and the fitness of the best individual for run 2 of the experiment summarized in Table 2, column 4 (block stacking).

As shown in the fourth column of Table 2, the probability of success for this problem is very high (0.70) despite using nine (out of 10) random initial configurations. It is worth noting that GP uses 167 fitness cases, cleverly constructed to cover the various classes of possible initial configurations [9]. Indeed, in real-life applications it is not always possible to predict the kind of cases that would make the system discover a solution. So, algorithms capable of generalizing well in face of random fitness cases are more advantageous.

### 6.4. Evolving cellular automata rules for the density-classification problem

Cellular automata (CA) have been studied widely as they are idealized versions of massively parallel, decentralized computing systems capable of emergent behaviors. These complex behaviors result from the simultaneous execution of simple rules at multiple local sites. In the density-classification task, a simple rule involving a small neighborhood and operating simultaneously in all the cells of a one-dimensional cellular automaton, should be capable of making the CA converge into a state of all 1s if the initial configuration (IC) has a higher density of 1s, or into a state of all 0s if the IC has a higher density of 0s.

The ability of GAs to evolve CA rules for the density-classification problem was intensively investigated [12-15], but the rules discovered by the GA performed poorly and were far from approaching the accuracy of the GKL rule, a human-written rule. GP was also used to evolve rules for the density-classification task [16], and a rule was discovered that surpassed the GKL rule and other human-written rules.

This section shows how GEP is successfully applied to this difficult problem. The rules evolved by GEP have accuracy levels of 82.513% and 82.55%, thus exceeding all human-written rules and the rule evolved by GP.

#### 6.4.1. The density-classification task

The simplest CA is a wrap-around array of $N$ binary-state cells, where each cell is connected to $r$ neighbors from both sides. The state of each cell is updated by a defined rule. The rule is applied simultaneously in all the cells, and the process is iterated for $t$ time steps.

In the most frequently studied version of this problem, $N=149$ and the neighborhood is 7 (the central cell is represented by "u"; the $r = 3$ cells to the left are represented by "c", "b", and "a"; the $r = 3$ cells to the right are represented by "1", "2", and "3"). Thus the size of the rule space to search for this problem is the huge number of $2^{128}$. Figure 14 shows a CA with $N = 11$ and the updated state for the cellular automaton "u" upon application of a certain transition rule.

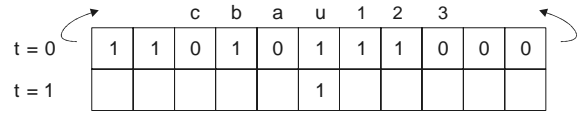

**Figure 14.** A one-dimensional, binary-state, $r = 3$ cellular automaton with $N = 11$. The arrows represent the periodic boundary conditions. The updated state is shown only for the central cell. The symbols used to represent the neighborhood are also shown.

The task of density-classification consists of correctly determining whether ICs contain a majority of 1s or a majority of 0s, by making the system converge, respectively, to an all 1s state (black or "on" cells in a space-time diagram), and to a state of all 0s (white or "off" cells). Because the density of an IC is a function of $N$ arguments, the actions of local cells with limited information and communication must be coordinated with one another to correctly classify the ICs. Indeed, to find rules that perform well is a challenge, and several algorithms were used to evolve better rules [14-17]. The best rules with performances of 86.0% (coevolution 2) and 85.1% (coevolution 1) were discovered using a coevolutionary approach between GA-evolved rules and ICs [17]. However, the aim of this section is to compare the performance of GEP with GAs and GP when applied to a difficult problem. And, in fact, GEP does evolve better rules than the GP rule, using computational resources that are more than four orders of magnitude smaller than those used by GP.

*6.4.2. Two gene expression programming discovered rules*

In one experiment $F = \{A, O, N, I\}$ ("A" represents the boolean function AND, "O" represents OR, "N" represents NOT, and "I" stands for IF) and $T = \{c, b, a, u, 1, 2, 3\}$. The parameters used per run are shown in Table 4, column 1.

**Table 4**
Parameters for the density-classification task.

|  | GEP$_1$ | GEP$_2$ |
|---|---|---|
| Number of generations | 50 | 50 |
| Population size | 30 | 50 |
| Number of ICs | 25 | 100 |
| Head length | 17 | 4 |
| Number of genes | 1 | 3 |
| Chromosome length | 52 | 39 |
| Mutation rate | 0.038 | 0.051 |
| 1-Point recombination rate | 0.5 | 0.7 |
| IS transposition rate | 0.2 | -- |
| IS elements length | 1,2,3 | -- |
| RIS transposition rate | 0.1 | -- |
| RIS elements length | 1,2,3 | -- |

The fitness was evaluated against a set of 25 unbiased ICs (i.e., ICs with equal probability of having a 1 or a 0 at each cell). In this case, the fitness is a function of the number of ICs $i$ for which the system stabilizes correctly to a configuration of all 0s or 1s after 2x$N$ time steps, and it was designed in order to privilege individuals capable of correctly classifying ICs both with a majority of 1s and 0s. Thus, if the system converged, in all cases, indiscriminately to a configuration of 1s or 0s, only one fitness point was attributed. If, in some cases, the system correctly converged either to a configuration of 0s or 1s, $f = 2$. In addition, rules converging to an alternated pattern of all 1s and all 0s were eliminated, as they are easily discovered and invade the populations impeding the discovery of good rules. And finally, when an individual program could correctly classify ICs both with majorities of 1s and 0s, a bonus equal to the number of ICs $C$ was added to the number of correctly classified ICs, being in this case $f = i + C$. For instance, if a program correctly classified two ICs, one with a majority of 1s and another with a majority of 0s, it receives 2+25=27 fitness points.

In this experiment a total of 7 runs were made. In generation 27 of run 5, an individual evolved with fitness 44:

```
0123456789012345678901234567890123456789012345678901
OAIIAucONObAbIANIb1u23u3a12aacb3bc21aa2baabc3bccuc13
```

Note that the ORF ends at position 28. This program has an accuracy of 0.82513 tested over 100,000 unbiased ICs in a 149x298 lattice, thus better than the 0.824 of the GP rule tested in a 149x320 lattice [16, 17]. The rule table of this rule (GEP$_1$) is shown in Table 5. Figure 15 shows three space-time diagrams for this new rule.

As a comparison, GP used populations of 51,200 individuals and 1000 ICs for 51 generations [16], thus a total of $51,200 \cdot 1,000 \cdot 51 = 2,611,200,000$ fitness evaluations were made, whereas GEP only made $30 \cdot 25 \cdot 50 = 37,500$ fitness evaluations. Therefore, in this problem, GEP outperforms GP by more than four orders of magnitude (69,632 times).

In another experiment a rule slightly better than GEP$_1$, with an accuracy of 0.8255, was obtained. Again, its performance was determined over 100,000 unbiased ICs in a 149x298 lattice. In this case $F = \{I, M\}$ ("I" stands for IF, and "M" represents the majority function with three arguments), and $T$ was obviously the same. In this case, a total of 100 unbiased ICs and three-genic chromosomes with sub-ETs linked by IF were used. The parameters used per run are shown in the second column of Table 4.

The fitness function was slightly modified by introducing a ranking system, where individuals capable of correctly classifying between 2 and 3/4 of the ICs receive one bonus equal to $C$; if between 3/4 and 17/20 of the ICs are correctly classified two bonus $C$; and if more than 17/20 of the ICs are correctly classified three bonus $C$. Also, in this experiment, individuals capable of correctly classifying only one kind of situation, although not indiscriminately, were differentiated and had a fitness equal to $i$.

**Table 5**
Description of the two new rules (GEP$_1$ and GEP$_2$) discovered using GEP for the density-classification problem. The GP rule is also shown. The output bits are given in lexicographic order starting with 0000000 and finishing with 1111111.

```
GEP₁    00010001 00000000 01010101 00000000 00010001 00001111 01010101 00001111
        00010001 11111111 01010101 11111111 00010001 11111111 01010101 11111111

GEP₂    00000000 01010101 00000000 01110111 00000000 01010101 00000000 01110111
        00001111 01010101 00001111 01110111 11111111 01010101 11111111 01110111

GP rule 00000101 00000000 01010101 00000101 00000101 00000000 01010101 00000101
        01010101 11111111 01010101 11111111 01010101 11111111 01010101 11111111
```

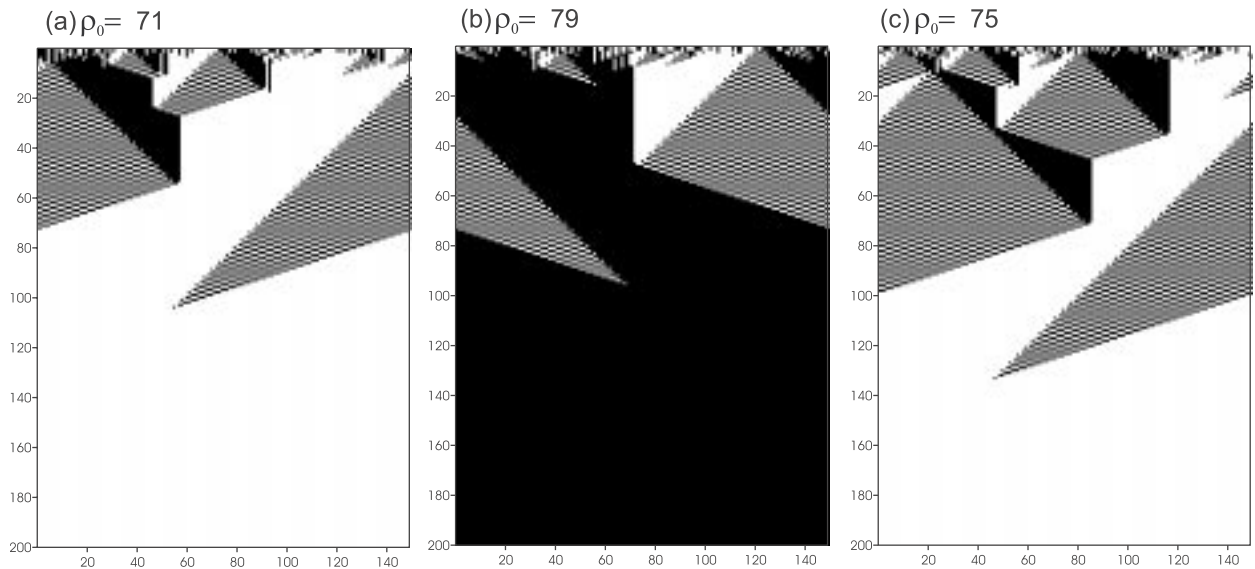

**Figure 15.** Three space-time diagrams describing the evolution of CA states for the $GEP_1$ rule. The number of 1s in the IC ($\rho_0$) is shown above each diagram. In **(a)** and **(b)** the CA correctly converged to a uniform pattern; in **(c)** it converged wrongly to a uniform pattern.

By generation 43 of run 10, an individual evolved with fitness 393:

```
012345678901201234567890120123456789012
MIuua1113b21cMIM3au3b2233bM1MIacc1cb1aa
```

Its rule table is shown in Table 5. Figure 16 shows three space-time diagrams for this new rule ($GEP_2$). Again, in this case the comparison with GP shows that GEP outperforms GP by a factor of 10,444.

### 6.5. Boolean concept learning

The GP rule and the 11-multiplexer are, respectively, boolean functions of seven and 11 activities. Whereas the solution for the 11-multiplexer is a well-known boolean function, the solution of the GP rule is practically unknown, as the program evolved by GP [16] is so complicated that it is impossible to know what the program really does.

This section shows how GEP can be efficiently applied to evolve boolean expressions of several arguments. Furthermore, the structural organization of the chromosomes used to evolve solutions for the 11-multiplexer is an example of a very simple organization that can be used to efficiently solve certain problems. For example, this organization (one-element genes linked by IF) was successfully used to evolve CA rules for the density-classification problem, discovering better rules than the GKL rule (results not shown).

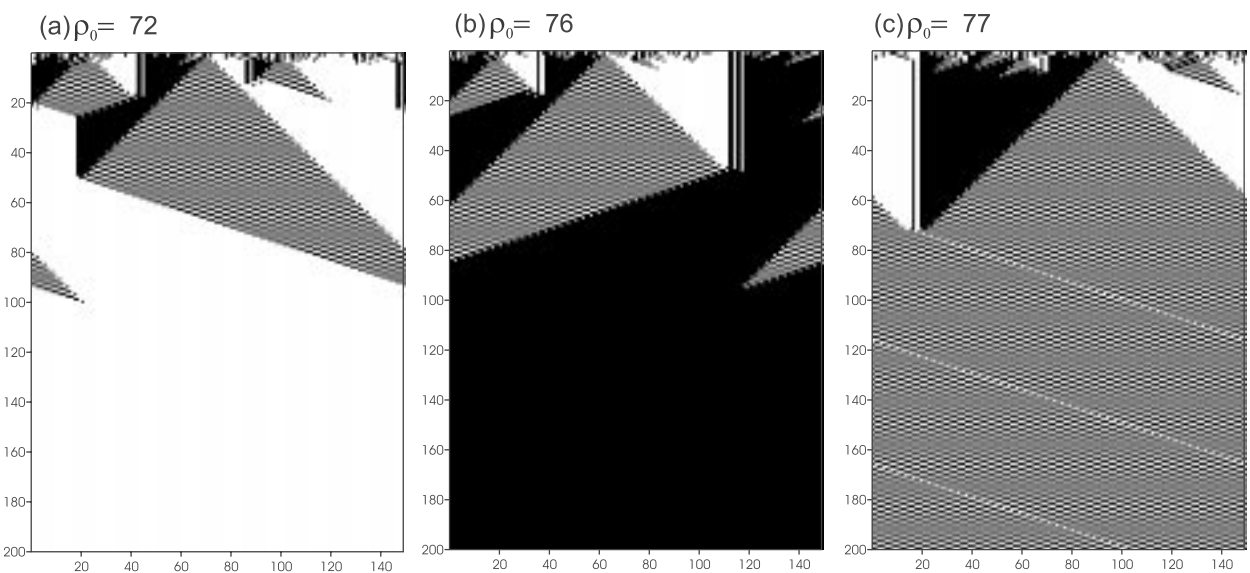

**Figure 16.** Three space-time diagrams describing the evolution of CA states for the $GEP_2$ rule. The number of 1s in the IC ($\rho_0$) is shown above each diagram. In **(a)** and **(b)** the CA converges, respectively, to the correct configuration of all 0s and all 1s; in **(c)** the CA could not converge to a uniform pattern.

*6.5.1. The genetic programming rule problem*

For this problem $F$ = {N, A, O, X, D, R, I, M} (representing, respectively: NOT, AND, OR, XOR, NAND, NOR, IF, and Majority, the first being a function of one argument, the second through fifth are functions of two arguments, and the last two are functions of three arguments), and $T$ = {c, b, a, u, 1, 2, 3}. The rule table ($2^7$=128 fitness cases) is shown in Table 5 and the fitness was evaluated by equation (4.2). Thus, $f_{max}$ = 128.

Three different solutions were discovered in one experiment:

```
MA3OOAMOAuOMRa1cc3cubcc2cu11ba2aacb331ua122uu1
X3RRMIMODIAIAAI3cauuc313bub2uc33ca12u233c22bcb
MMOIOcXOMa3AXAu3cc112ucbb3331uac3cu3auubuu2ab1
```

Careful analysis of these programs shows that the GP rule is, like the GKL rule, a function of five arguments: c, a, u, 1, and 3.

*6.5.2. The 11-multiplexer problem*

The task of the 11-bit boolean multiplexer is to decode a 3-bit binary address (000, 001, 010, 011, 100, 101, 110, 111) and return the value of the corresponding data register ($d_0$, $d_1$, $d_2$, $d_3$, $d_4$, $d_5$, $d_6$, $d_7$). Thus, the boolean 11-multiplexer is a function of 11 arguments: three, $a_0$ to $a_2$, determine the address, and eight, $d_0$ to $d_7$, determine the answer. As GEP uses single-character chromosomes, $T$ = {a, b, c, 1, 2, 3, 4, 5, 6, 7, 8} which correspond, respectively, to {$a_0$, $a_1$, $a_2$, $d_0$, $d_1$, $d_2$, $d_3$, $d_4$, $d_5$, $d_6$, $d_7$}.

There are $2^{11}$ = 2048 possible combinations for the 11 arguments of the boolean 11-multiplexer function. For this problem a random sampling of the 2048 combinations was used each generation as the fitness cases for evaluating fitness. The fitness cases were assembled by address, and for each address a sub-set of 20 random combinations was used each generation. Therefore, a total of 160 random fitness cases were used each generation as the adaptation environment. In this case, the fitness of a program is the number of fitness cases for which the boolean value returned is correct, plus a bonus of 180 fitness points for each sub-set of combinations solved correctly as a whole. Therefore, a total of 200 fitness points was attributed for each correctly decoded address, being the maximum fitness 1600. The idea was to make the algorithm decode one address at a time. And, in fact, the individuals learn to decode first one address, then another, until the last one (see Figure 17).

To solve this problem, multigenic chromosomes composed of 27 genes were used, each gene consisting of only one terminal. Thus, no functions were used to generate the chromosomes, although the sub-ETs were posttranslationally linked by IF.

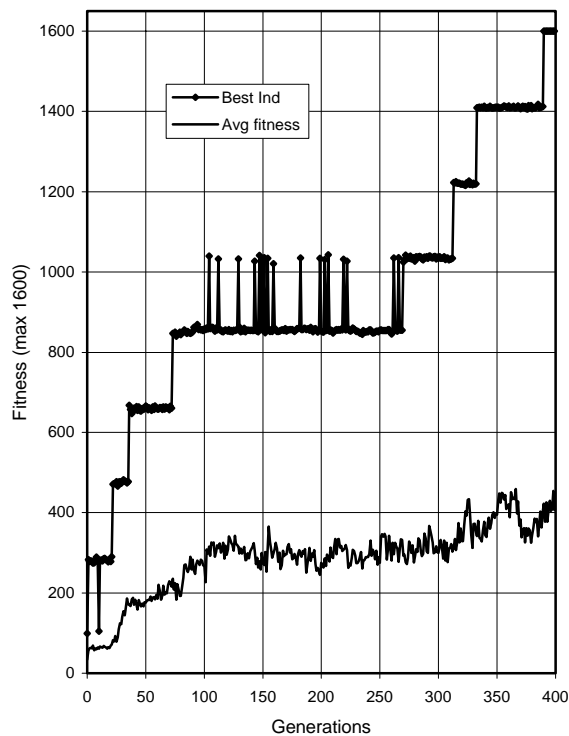

**Figure 17.** Progression of average fitness of the population and the fitness of the best individual for run 1 of the experiment summarized in Table 2, column 5 (11-multiplexer).

The parameters used per run are shown in Table 2, column 5. The first correct solution in this experiment was found in generation 390 of run 1 (the characters are linked 3 by 3, forming an ET with depth 4, composed of 40 nodes, the first 14 nodes being IFs, and the remaining nodes, the chromosome characters; see K-expression (3.12) and Figure 5):

```
3652bb5bbba4c87c43bcca62a51
```

which is a universal solution for the 11-multiplexer. Figure 17 shows the progression of average fitness of the population and the fitness of the best individual for run 1 of the experiment summarized in Table 2, column 5.

As shown in the fifth column of Table 2, GEP solves the 11-multiplexer with a success rate of 0.57. It is worth noting that GP could not solve the 11-multiplexer with a population size 500 for 51 generations [18], and could only solve it using 4,000 individuals [9].

## 7. Conclusions

The details of implementation of gene expression programming were thoroughly explained allowing other researchers to implement this new algorithm. Furthermore, the problems chosen to illustrate the functioning of GEP show that the new paradigm can be used to solve several problems from different fields with the advantage of running efficiently in a personal computer. The new concept behind the linear chromosomes and the ETs enabled GEP to considerably outperform existing adaptive algorithms. Therefore, GEP offers new possibilities for solving more complex technological and scientific problems. Also important and original is the multigenic organization of GEP chromosomes, which makes GEP a truly hierarchical discovery technique. And finally, gene expression algorithms represent nature more faithfully, and therefore can be used as computer models of natural evolutionary processes.

## Acknowledgments

I am very grateful to José Simas for helping with hardware problems, for reading and commenting on the manuscript, and for his enthusiasm and support while I was grasping the basic ideas and concepts of GEP.